# Reconfigurable Manipulator Simulation for Robotics and Multimodal Machine Learning Application: Aaria

Arttu Hautakoski, Mohammad M. Aref, and Jouni Mattila
*Laboratory of Automation and Hydraulic Engineering*
*Tampere University of Technology*
Tampere, Finland

*Abstract*—This paper represents a systematic way for generation of Aaria, a simulated model for serial manipulators for the purpose of kinematic or dynamic analysis with a vast variety of structures based on Simulink SimMechanics. The proposed model can receive configuration parameters, for instance in accordance with modified Denavit-Hartenberg convention, or trajectories for its base or joints for structures with 1 to 6 degrees of freedom (DOF). The manipulator is equipped with artificial joint sensors as well as simulated Inertial Measurement Units (IMUs) on each link. The simulation output can be positions, velocities, torques, in the joint space or IMU outputs; angular velocity, linear acceleration, tool coordinates with respect to the inertial frame. This simulation model is a source of a dataset for virtual multimodal sensory data for automation of robot modeling and control designed for machine learning and deep learning approaches based on big data. [1]

I. INTRODUCTION

In this paper, we present a reconfigurable manipulator simulator for robotic and multimodal machine learning applications. We present the model structure and its components, model's inputs and outputs and some sample results generated by the model. The original goal was to gather sensor data from a variety of manipulators. Collecting a large and diverse enough dataset from physical manipulators was not feasible and neither was building a new simulator for every simulation. The solution for these problems was to build a reconfigurable simulator capable of changing its dimensions and structure simply by changing its parameters. In addition to using the simulator for its original purpose, it can also be used for applications that have specific requirements by tuning the parameters. The model will be made publicly available in the future.

*A. Model*

The model is equipped with several modelled Inertial Measurement Units (IMUs) to measure and record data for machine learning applications. The model consists of a floating base, tool and up to six modular joint systems. The simulator can build and simulate different kind of manipulators based on Denavit-Hartenberg parameters using the modified notation and parameters for the joint types and trajectories for the base and joints. Each modular joint's type can be configured to be revolute, prismatic or empty. The model was built in MATLAB Simulink environment using Simscape Multibody blocks. The reconfigurability was implemented by using Variant Subsystem blocks and selecting a certain joint model based on input parameters. The coordinate transformations and sensor modelling was done with Simscape blocks.

*B. Literature review*

Reconfigurable manipulators in MATLAB environment have been explored before (Corke, 1996). However, the ability to place sensors on the simulated manipulator is not supported. Even a physical reconfigurable manipulator has been researched (Kereluk and Emami, 2015) A physical reconfigurable manipulator is viable for obtaining a variety of sensor data but it lacks the capability to produce a large enough dataset in reasonable amount of time. Datasets have been obtained before by simulation (Vijayan et al, 2013) (Veres et al, 2017) but the simulators have been built for a specific purpose and the datasets are only suitable for the application they were generated for.

*C. Background information*

The manipulator is defined by modified Devavit-Hartenberg (DH) parameters (Craig, 2005). Each transformation from a joint to next is described by four parameters. Parameter a, defines the length of the link from current joint to the next one along x-axis. Parameter α, defines the rotation of the joint around x-axis. Parameter d, defines the distance along the z-axis to the next joint. Parameter θ, defines the rotation around z-axis. For revolute joints, the θ parameter would define the angle of the joint so that is not used to define the joint but measured instead. Similarly the parameter d, would define the prismatic joint's position so that is the measured value of the joint.

Real physical manipulator's joint values can be measured with joint encoders. In the case of the model, the values are measured with the outputs from the joint blocks.

II. MODEL DESCRIPTIONS

The model starts from a World Frame-block that defines the coordinate axes of the building environment. Solver Configuration and Mechanism Configuration blocks are attached to it and they define the solver settings and the gravity. The World Frame connects to the Floating base to determine its location. Modular joints are connected to the base module as a chain that ends in the tool module.

---

[1] The model is will be publicly available by contacting the corresponding author: m.aref [at] ieee [dot] org

### A. Base

The Floating base has the ability to move along any axis as well as rotate around any axis. In the model, this is implemented by connecting a Cartesian Joint-block to the World Frame. The user provides the inputs for Cartesian movement, like every other movement, as parameters for a Sine Wave block. The parameters for the sine wave are amplitude, bias, frequency and phase. The resulting signal is then converted into a physical signal so it can be used with Simscape blocks. The conversion uses the original signal and its two derivatives. Each Cartesian axis has its own input signal. The Rotation of the base is done by using a Gimbal Joint-block, which is attached as the follower of the Cartesian Joint-block. The rotation signals around each axis are provided in the same manner as with the Cartesian joint. The signal coming out of the Gimbal Joint-block carries the information of the current rotation and translation in relation to the world frame. A single IMU-sensor is connected to this signal and the world frame signal to track and record the positon and orientation of the base.

### B. IMU

The movements of the manipulator joints are measured with modelled IMUs. The model of the IMU is based on Analog Devices ADIS16485 inertial sensor (Analog Devices, 2018). The model is made by using Simscape Transform sensor-blocks. The block takes the World Frame and the current frame and calculates the angular velocity and the acceleration of the sensor. Each axis has its own signal so one sensor will output six signals. Some measurement noise is added to the signals to achieve output that is more realistic. The Sensor position and orientation can be changed by altering the parameter file. In the current model, the sensors are placed to the middle and near the end of each link at the opposite sides of the link so they do not share a same axis. Special cases are the base and the tool where only one sensor is placed at the bottom of the base and at the tip of the tool.

### C. Modular joint

The modular joint block includes a model of a prismatic joint, revolute joint and an empty link. The desired joint type can be selected by setting the variant subsystem parameters in the parameter file. The model is built inside a masked subsystem. The mask asks the user to input the link number. In this case, the modular joints are numbered 1-6. The tool's link number is 7 and the base's link number is 8. The link number parameters is used to pick specific parameters from the parameter file for each link. Under the mask, there is a variant subsystem block. Variant subsystem is a block that can include multiple subsystems. Only one of the subsystems inside the variant subsystem can be active at a time. The active variant is controlled by the variant control parameter and its associated condition. In this model, there are three different subsystems which are revolute joint, prismatic joint and an empty joint. The model layout is presented in figure 1.

The inputs to all the subsystems are the signal with the information about the translation and rotation of the frame and the reference signal coming from the original world frame. The outputs are measurement data from the attached IMUs, joint rotation or translation measurement, depending on the link type, and the current rotation and translation of the system.

### D. Revolute joint

The goal was to make a reconfigurable system based on DH-parameters. The DH-coordinate system is different to what Simscape uses, so a rotation needs to be performed before doing any operations related to those frames. In order to perform rotations and translations according to the DH-parameters, the coordinate frame needs to be transformed into the DH coordinate system. This is done by rotating –pi/2 radians around y-axis and pi/2 around x-axis. In the front view of the model, the world frame x-axis points to the right, y-axis forward and the z-axis points up. After the first rotation to DH-frame, x-axis is pointing up, y-axis to the left and z-axis towards the viewer. A coordinate frame is drawn on the visualization at this point. The next operation is α rotation based on the DH-parameters. This is done by using a Rigid Transform block that is set to rotate according to a rotation matrix specified in the parameter file. The rotation happens about the x-axis. Next operation is the translation along the z-axis. This is done by using another Rigid Transform block. This time, the translation is simply parametrized by the corresponding DH-parameter d, of that link. The joint's visualization is drawn as a cylinder by using the Solid-block. Simscape draws objects by setting the current position as the center point and aligning the object towards the z-axis. Next block is the actual Revolute Joint block that rotates around the z-axis. The rotation is controlled by a sine wave shaped motion input specified in the parameter file. The motion input signal consists of the input and its first two derivatives. The joint senses its position and that signal is routed out of the subsystem and recorded. After the joint block, the IMU subsystem is connected to the model to sense the current rotation of the frame. The last operation of revolute joint's DH-parameter transforms, is the translation along the x-axis by the amount defined by the DH-parameter a. A link needs to be drawn between the current joint and the next one. The link is oriented along the x-axis and Simscape draws objects along the z-axis so this time the link cannot be drawn as simply as the joint. First, a counter rotation is performed to reverse the rotation that was done when transitioning into the DH-coordinate frame. Now the z-axis is pointing towards the next link like DH-coordinate frame's x-axis does. Next, a translation is performed to move the coordinate frame half of *a* parameter's value along the new z-axis to get to halfway point of the link. At this point, the link is drawn by using a brick shaped Solid-block parametrized by the parameter file. Another translation is performed to move the current coordinate frame from the middle of the link to the end of the link. Before ending the model, the coordinate frame is rotated to the DH-coordinates, a coordinate frame is drawn and the frame is rotated back out of the DH-coordinate frame once more. Translations in revolute joint are presented in table 1.

Table 1. Transformations in a revolute joint model.

### E. Prismatic joint

The prismatic joint is one of the options inside the variant subsystem. The prismatic joint model shares many similarities with the revolute joint model. The rotation to DH-coordinate frame, drawing the coordinate frame visualization and the α rotation are done exactly like in the revolute joint model. After that, instead of doing the d translation, a θ rotation is performed about the z-axis by the amount specified in the DH-parameters. At this point, the rail that the prismatic joint moves on, is drawn. The length of the rail is specified by the d-parameter. After that, the Prismatic Joint-block is placed into the model. The configurations of the prismatic joint's inputs and outputs are the same as with the revolute joint. The only difference is that the prismatic joint moves along the z-axis instead of rotating around it. The rest of the model is the same as in the revolute joint model, including IMU placement, rotations between the coordinate frames, translation according to the "a" parameter and drawing the links and coordinate frames. Translations in the prismatic joint are presented in table 2.

Table 2. Transformations in a prismatic joint model.

| Matrix | Description | Value |
|---|---|---|
| ${^{\{W\}}}T_{\{DH\}}$ | Rotation from Simscape frame to DH | $\begin{bmatrix} 0 & -1 & 0 & 0 \\ 0 & 0 & -1 & 0 \\ 1 & 0 & 0 & 0 \\ 0 & 0 & 0 & 1 \end{bmatrix}$ |
| ${^{\{DHi-1\}}}T_{\{DHi\}}$ | Rotation about x-axis | $\begin{bmatrix} 1 & 0 & 0 & 0 \\ 0 & \cos(\alpha) & -\sin(\alpha) & 0 \\ 0 & \sin(\alpha) & \cos(\alpha) & 0 \\ 0 & 0 & 0 & 1 \end{bmatrix}$ |
| ${^{\{DHi\}}}T_{\{DHth\}}$ | Rotation about z-axis | $\begin{bmatrix} \cos(\theta) & -\cos(\theta) & 0 & 0 \\ \cos(\theta) & \cos(\theta) & 0 & 0 \\ 0 & 0 & 1 & 0 \\ 0 & 0 & 0 & 1 \end{bmatrix}$ |
| ${^{\{DHth\}}}T_{\{DHi+1\}}$ | Translation along x-axis | $\begin{bmatrix} 1 & 0 & 0 & a \\ 0 & 1 & 0 & 0 \\ 0 & 0 & 1 & 0 \\ 0 & 0 & 0 & 1 \end{bmatrix}$ |
| ${^{\{DH\}}}T_{\{W\}}$ | Rotation from DH frame to Simscape | $\begin{bmatrix} 0 & 0 & 1 & 0 \\ -1 & 0 & 0 & 0 \\ 0 & -1 & 0 & 0 \\ 0 & 0 & 0 & 1 \end{bmatrix}$ |

### F. Empty joint

The third link type in the variant subsystem is an empty link. This system has the same inputs and outputs as the other link types. The empty link does not perform any transformations. It simply passes the current transform signal through without changing it in any way. Simulink requires the reference coordinate input to the subsystem but it is not used for

| Matrix | Description | Value |
|---|---|---|
| ${^{\{W\}}}T_{\{DH\}}$ | Rotation from Simscape frame to DH | $\begin{bmatrix} 0 & -1 & 0 & 0 \\ 0 & 0 & -1 & 0 \\ 1 & 0 & 0 & 0 \\ 0 & 0 & 0 & 1 \end{bmatrix}$ |
| ${^{\{DHi-1\}}}T_{\{DHi\}}$ | Rotation about x-axis | $\begin{bmatrix} 1 & 0 & 0 & 0 \\ 0 & \cos(\alpha) & -\sin(\alpha) & 0 \\ 0 & \sin(\alpha) & \cos(\alpha) & 0 \\ 0 & 0 & 0 & 1 \end{bmatrix}$ |
| ${^{\{DHi\}}}T_{\{DHid\}}$ | Translation along z-axis | $\begin{bmatrix} 1 & 0 & 0 & 0 \\ 0 & 1 & 0 & 0 \\ 0 & 0 & 1 & d \\ 0 & 0 & 0 & 1 \end{bmatrix}$ |
| ${^{\{DHid\}}}T_{\{DHi+1\}}$ | Translation along x-axis | $\begin{bmatrix} 1 & 0 & 0 & a \\ 0 & 1 & 0 & 0 \\ 0 & 0 & 1 & 0 \\ 0 & 0 & 0 & 1 \end{bmatrix}$ |
| ${^{\{DH\}}}T_{\{W\}}$ | Rotation from DH frame to Simscape | $\begin{bmatrix} 0 & 0 & 1 & 0 \\ -1 & 0 & 0 & 0 \\ 0 & -1 & 0 & 0 \\ 0 & 0 & 0 & 1 \end{bmatrix}$ |

anything. The empty link has no visualizations and no sensors. Even though the model does not have sensors, it produces placeholder sensor data. Instead of acceleration and angular velocity data, the empty link outputs a 12-signal bus of constant zero to match the output dimensions of a non-empty link with two IMUs. The position measurement is handled by a single constant zero output signal. An active empty link model will show up in the measurement data as a column of zeros at the joint values and as 12 consecutive columns of zeros at the sensor data.

### G. Tool

The tool takes the current transformation data and reference coordinates as inputs and it outputs a bus of six signals that comes from the IMU placed at the tip of the tool. The tool is not a joint but the model has its own link number and DH-parameter table. The *a* parameter from DH-table is used to determine the length of the tool and the link number is used to assign that length to the tool as well as the sensor parameters. The tool model starts by translation along the z-axis by half the amount of DH-parameter a. Next, the tool is drawn like any other link and the second half of the translation is performed to move the current coordinate frame to the end of the tool. A rotation to the DH-coordinate frame is performed in the same block with the second translation. A DH-coordinate frame is drawn and an IMU is placed at the tip of the tool.

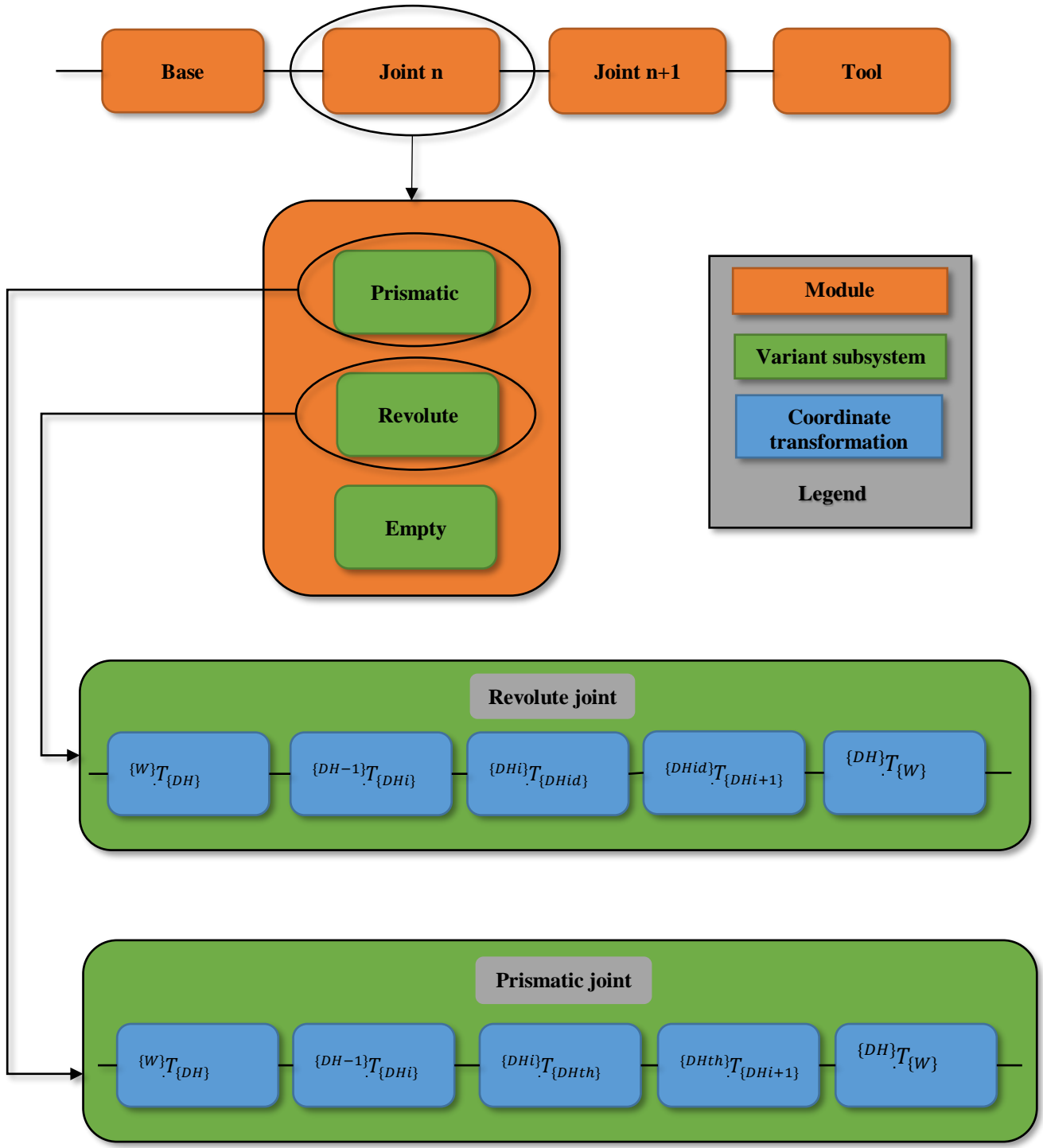

*Figure 1. Model layout and transformations within joint models.*

### H. Callback functions

The top level of the model has an empty subsystem block named "My_commands". The block includes the callback functions the model. The model has the capability to initialize itself when the model is loaded and after every simulation. If the model is commanded to run before it is loaded, the LoadFcn function will trigger and it will check the workspace for a variable called "mfiledone". If the variable does not exist, the model will initialize itself by running the parameter file and the simulation is executed. However, if the variable does exist, the simulation is executed without initialization.

After the simulation, StopFcn will trigger and it will check the value of "randomize" parameter. If the value is 1, the model will run the parameter file. Otherwise no operations are done at the end of the simulation.

## III. INPUT/OUTPUT PROTOCOLS

### A. Inputs

The inputs of the model are provided in three arrays: extended DH-parameter table, joint oscillation table and base oscillation table.

#### 1) DH-parameter table

The DH-parameter table has five columns and eight rows. The values are in the columns as follows: $\alpha$, $a$, $\theta$, $d$, link type. Parameters $\alpha$ and $\theta$ angles are measured in radians. Parameters $a$ and $d$ are measured in meters. Link types are defined as follows: zero means empty link, one means revolute joint and two means prismatic joint. Each row in the array defines the parameters for one joint. The first six rows are assigned to the modular links, seventh row to the tool and eighth to the base. Although most of the DH-parameters for the base and tool are unused, some values are still required to keep the array dimensions consistent.

#### 2) Joint oscillation table

The joint oscillation is implemented with Sine Wave blocks. The oscillation data is parameters for those blocks. The joint oscillation data is provided in 8x5 array where the columns have the following values: amplitude, prismatic amplitude, bias, frequency and phase. The values of the first six values are assigned to the modular links in order and the last two rows are unused. Revolute joints use the amplitude value and prismatic joints use the prismatic amplitude value.

#### 3) Base oscillation table

Base oscillation parameters are provided in a 2x4 array where the columns have values for amplitude, bias, frequency and phase. The first row defines the translation of the base and the second row defines the rotation.

#### 4) Randomization

The parameter file expects the input arrays to be named senSenDH, senSenJ and senSenB. The arrays can be manually created in MATLAB but they can also be generated automatically by the model by running it first with randomized values. The parameters can be randomized for each simulation by setting the "MODE" parameter as 1 and running the parameter file. This generates a random DH-parameter table as well as trajectories for base and joints. The parameters will be randomized automatically after each simulation if the "randomize" parameter is set to 1. Setting the "MODE" variable to 2 instructs the model to read the parameters from the arrays in the workspace. MATLAB will give a warning if the parameter file is run in MODE 2 without proper input arrays in the workspace. The model was designed to be used with randomized values and running the parameter file in any non-randomized way will produce a notification about the randomization being turned off.

### B. Outputs

All the simulation measurement data is collected into a single array named senSen. The order of the signals in the array is time, joint values, base sensor, tool sensor and the sensor data of each link from first to last. Time and joint values are all one column each. Base and tool have only one IMU each so those signals are both six columns wide. Each link has two IMUs so the sensor data from each link is 12 columns. Data from empty links shows up as zeros. In addition to measurement data, the input data is collected as well. The used DH-parameter tables and the trajectories for joints and the base are saved in their own arrays for later reference. The saved input data is saved in the same array form and named senSenDH, senSenJ and senSenB respectively. These arrays can be edited and the changes will be visible during the next simulation if the parameters are not set to randomize.

## IV. SAMPLE RESULTS

The built model and simulation can be viewed in MATLAB's Mechanics explorer. Revolute joints are visualized as green cylinders and prismatic joints as green rods. Links between the joints are visualized as blue bricks. The starting point of the manipulator is at the world frame which is visualized as a green coordinate frame. The manipulator ends at the tooltip which has a yellow coordinate frame. The red coordinate frames represent the current state of the DH-coordinates. The red DH-coordinate frames are drawn at the end of each link and not on the joint itself. The closest coordinate frame to the joint does not represent that joint's orientation. The tool is visualized as a blue brick just like normal links. The base has no visualization. The movement of the base can be seen as the first link's movement in relation to the world frame. Sensors are visualized as purple coordinate frames. The size of the sensor frames have been reduced to make the DH-coordinate frames easier to see.

Pictures of example manipulator configurations and their visualizations are presented below. The pictures are accompanied with a DH-parameter table that was used as an input for the model. The grey label column and row are not part of the input. The link lengths and rotations can be anything but in this demonstration, link lengths are between 0.8 and 0.9 meters and the rotations van be positive or negative 90 degrees of zero.

The first example is presented in figure 2 and the parameters it was generated from are presented in table 3. It is a six jointed manipulator with two prismatic joints and four revolute joints. The first transformation from world frame to DH-coordinate frame can be seen as partially overlapping green and red coordinate frames.

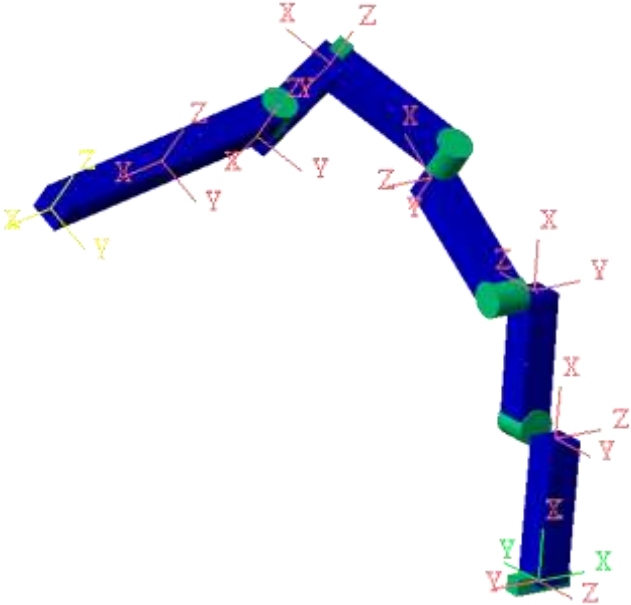

*Figure 2. Example manipulator.*

*Table 3. DH-parameter table for example manipulator in figure 2.*

| Link | α | a | θ | d | Link type |
|---|---|---|---|---|---|
| 1 | π/2 | 0.89 | 0 | 0.14 | 2 |
| 2 | π/2 | 0.83 | π/2 | 0.29 | 1 |
| 3 | π/2 | 0.88 | 0 | 0.22 | 1 |
| 4 | π/2 | 0.85 | π/2 | 0.24 | 1 |
| 5 | 0 | 0.86 | π/2 | 0.13 | 2 |
| 6 | 0 | 0.89 | 0 | 0.23 | 1 |
| 7 (Tool) | π/2 | 0.89 | π/2 | 0.12 | 1 |
| 8 (Base) | - π/2 | 0.86 | 0 | 0.29 | 1 |

The second example is presented in figure 3 and the parameters it was generated from are presented in table 4. This manipulator has three prismatic joints and two revolute joints. The link type of the second joint is empty. This means that the second row of parameters is unused and any parameters on that row have no effect on the configuration. In addition, revolute joints do not use the θ parameter as input for the joint angle. The joint angle is initialized as zero and the angle during the simulation is determined by the joint oscillation trajectory. Similarly, the d parameter does not determine the position of prismatic joints but it is used as a value for the range of the joint.

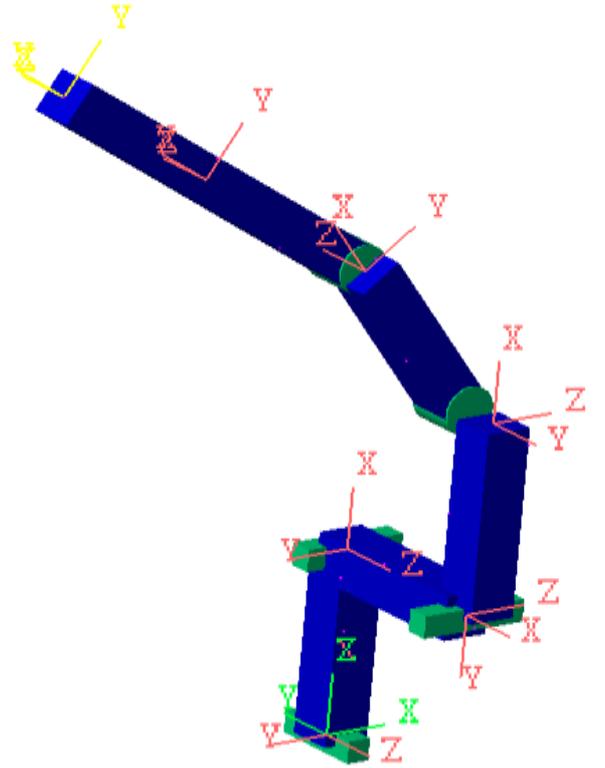

*Figure 3. Second example manipulator.*

*Table 4. DH-parameters table for example manipulator in figure 3.*

| Link | α | a | θ | d | Link type |
|---|---|---|---|---|---|
| 1 | 0 | 0.84 | 0 | 0.17 | 2 |
| 2 | 0 | 0.82 | π/2 | 0.26 | 0 |
| 3 | π/2 | 0.82 | π/2 | 0.21 | 2 |
| 4 | 0 | 0.85 | - π/2 | 0.23 | 2 |
| 5 | π/2 | 0.83 | -π/2 | 0.26 | 1 |
| 6 | 0 | 0.87 | 0 | 0.12 | 1 |
| 7 (Tool) | π/2 | 0.88 | π/2 | 0.27 | 1 |
| 8 (Base) | - π/2 | 0.85 | π/2 | 0.20 | 1 |

V. CHALLENGES AND GOALS

*A. Challenges*

Variant subsystems offer great flexibility for modelling complex systems. However, the variant subsystem blocks themselves have some limitations. The variant control variables and conditions do not support parametrization so each variable has to be manually renamed after duplicating a module block. Every other parameter in the module blocks is parametrized by the link number given in the mask at the top level of the module.

The variant subsystems also require the variant control parameters unusually early in the simulation compared to other blocks. Automatically initializing these parameters is challenging because the only callback function executed early enough to set these parameters is LoadFcn. The problem with using LoadFcn is that it only runs when the model is first loaded. The model has to be closed and loaded again to run the callback function repeatedly.

Alternatively, the parameters can be initialized at the end of the simulation so there is no need to reinitialize at the start of the simulation. This solution is effective for repeated manual simulations but causes difficulties when trying to run automatic parallel simulations.

### B. Goals

The model is still a work in progress. The goal is to overcome the previously presented challenges and improve the user experience. In addition to general improvements, the model will be modified to allow torque and force inputs for the joints in addition to the motion trajectories. This change will be accompanied by improved inertia model based on link dimensions and density.